\documentclass[runningheads]{llncs}
\usepackage[T1]{fontenc}
\usepackage{times}
\usepackage{soul}
\usepackage{url}
\usepackage[hidelinks]{hyperref}
\usepackage[utf8]{inputenc}
\usepackage[small]{caption}
\usepackage{graphicx}
\usepackage{amsmath}
\usepackage{booktabs}
\usepackage{algorithm}
\usepackage{algorithmic}
\usepackage[switch]{lineno}
\usepackage{makecell}
\usepackage[edges]{forest}
\definecolor{hidden-draw}{RGB}{205, 44, 36}
\definecolor{hidden-blue}{RGB}{194,232,247}
\definecolor{hidden-orange}{RGB}{243,202,120}
\definecolor{hidden-yellow}{RGB}{242,244,193}
\definecolor{tree-level-1}{RGB}{245,20,85}
\definecolor{tree-level-2}{RGB}{246,86,118}
\definecolor{tree-level-3}{RGB}{248,177,193}
\definecolor{tree-leaf}{RGB}{176,230,198}

\begin{document}
\title{Toward Efficient Deep Spiking Neuron Networks:\\A Survey On Compression}

\author{Hui Xie\inst{1} \and
Ge Yang\inst{1} \and Wenjuan Gao\inst{1}}
\institute{Beihang University, Beijing, CHINA\\
\email{\{xiehui,20231102,wjgao\}@buaa.edu.cn}
}

% \author{Anonymous authors}
% \institute{Paper under double-blind review}

\maketitle
\begin{abstract}
With the rapid development of deep learning, Deep Spiking Neural Networks (DSNNs) have emerged as promising due to their unique spike event processing and asynchronous computation. When deployed on neuromorphic chips, DSNNs offer significant power advantages over Deep Artificial Neural Networks (DANNs) and eliminate time and energy consuming multiplications due to the binary nature of spikes (0 or 1). Additionally, DSNNs excel in processing temporal information, making them potentially superior for handling temporal data compared to DANNs. However, their deep network structure and numerous parameters result in high computational costs and energy consumption, limiting real-life deployment. To enhance DSNNs efficiency, researchers have adapted methods from DANNs, such as pruning, quantization, and knowledge distillation, and developed specific techniques like reducing spike firing and pruning time steps. While previous surveys have covered DSNNs algorithms, hardware deployment, and general overviews, focused research on DSNNs compression and efficiency has been lacking. This survey addresses this gap by concentrating on efficient DSNNs and their compression methods. It begins with an exploration of DSNNs' biological background and computational units, highlighting differences from DANNs. It then delves into various compression methods, including pruning, quantization, knowledge distillation, and reducing spike firing, and concludes with suggestions for future research directions.
\keywords{Deep Spiking Neuron Networks  \and Pruning \and Quantization \and Knowledge distillation \and Reducing Spiking Firing \and Compression}
\end{abstract}
\section{Introduction}
As early as 1997, Maass~\cite{Maass1996NetworksOS} classified Spiking Neural Networks (SNNs) as third-generation neural networks based on their computational units, specifically spiking neurons or "integrate-and-fire neurons" . In contrast, first-generation neural networks, developed around the 1950s and known as perceptrons, were based on McCulloch-Pitts neurons. These networks struggled with problems that were not linearly separable, such as the XOR operation. To address these limitations, second-generation neural networks, or Artificial Neural Networks (ANNs), were developed, utilizing activation functions that apply a continuous set of output values to a weighted sum (or polynomial) of the inputs.

With the advancement of neuromorphic computing~\cite{Roy2019TowardsSM}, SNNs have demonstrated significant potential due to their unique spike event processing and asynchronous computation capabilities . When deployed on neuromorphic chips, SNNs offer substantial power advantages over ANNs and avoid time and energy consuming multiplication operations due to the binary nature of spikes (0 or 1)~\cite{Bhattacharjee2023AreST} . Moreover, SNNs more closely mimic the complex workings of the biological brain~\cite{Huang2023DeepSN}, enhancing their potential to achieve genuine artificial intelligence . Their ability to process temporal information naturally makes them particularly suited for tasks involving temporal data~\cite{Li2023EfficientHA}, surpassing the capabilities of ANNs in this regard . SNNs are also uniquely suited for applications like event-based cameras~\cite{Gallego2019EventBasedVA}, which are event-driven, asynchronous, and binary, making them ideal for processing with SNNs on neuromorphic chips .

In recent years, deep learning~\cite{lecun2015dl} has achieved tremendous success with DANNs~\cite{chen2021imageClass,zou2023objectDetection,minaee2021imageSegment,yin2017nlp} and, correspondingly, with DSNNs~\cite{Kim2020SpikingBN,Zhu2022TrainingSNNs,Lv2023SpikingCNTextClass}. Numerous new architectures have been proposed to improve performance on various tasks in DANNs~\cite{krizhevsky2012alexNet,simonyan2014vggNet,he2016resNet,huang2017denseNet} , with similar advancements seen in DSNNs, including Spiking VGG~\cite{Lee2019spikingVgg,Sengupta2018spikingVgg} , Spiking ResNet~\cite{Hu2021SpikingResNet,Fang2021SewResNet} , and Spiking Transformers~\cite{Zhou2022Spikformer,Shi2024SpikingResformer,Zhang2024SGLFormer} .

However, the deep network structure and large number of parameters in these models lead to significant computational costs and energy consumption, limiting their practical deployment. To achieve efficient DSNNs, researchers have adapted methods from DANNs, such as pruning, quantization, and knowledge distillation. Additionally, unique methods specific to DSNNs, such as reducing spike firing and pruning time steps, have been developed.

Previous surveys on SNNs have primarily focused on algorithms~\cite{dampfhoffer2023snnBPSurvey,zhou2024directTraingSurvey} , hardware deployment~\cite{bouvier2019snnhardwareimpleSurvey} , and general overviews~\cite{pfeiffer2018dlSnnSurvey} , without specifically addressing DSNNs compression and efficiency. Conversely, there is extensive research on compressing DANNs, including structured pruning~\cite{he2024structured}, quantization~\cite{gholami2022surveyQuanti}, knowledge distillation~\cite{gou2021surveyknowledgeDistillation} . Therefore, a focused survey on DSNNs compression is needed.

This survey addresses this gap by concentrating on efficient DSNNs and their compression methods. It begins with an exploration of the biological background and computational units of SNNs to understand how they differ from ANNs. It then analyzes various compression methods, including pruning, quantization, knowledge distillation, and reducing spike firing. Finally, it suggests directions for future research.

\section{Background}
The introduction of activation functions in neural networks is primarily driven by the need to introduce non-linear functions into a linear system, thereby increasing its complexity and enhancing its representative capabilities. In contrast, the computational units in SNNs, such as the Leaky Integrate-and-Fire (LIF) neuron or Integrate-and-Fire (IF) neuron, are designed to more closely mimic the properties of biological neurons. This distinction in computational units is a fundamental difference between SNNs and ANNs.

\subsection{Biological Background}

A typical neuron consists of three main parts: dendrites, soma, and axon. Dendrites collect input signals from other neurons and transmit them to the soma. The soma acts as a computational unit, generating an action potential when the accumulated incoming current causes the membrane potential to exceed a certain threshold. This action potential then travels along the axon and transmits the signal to the next neuron through synapses at the axon's terminal.

\subsubsection{LIF Model.}

The Leaky Integrate-and-Fire (LIF) model, first proposed by Lapicque in 1907, describes the process of action potentials in neurons. Neurons fire impulses when the membrane potential reaches the threshold voltage $V_{threshold}$, after which the membrane potential resets to the resting potential  $V_{reset}$. The LIF model focuses on the patterns of sub-threshold potential voltage variations.~\cite{Dayan2001NeuronModel}.
\begin{equation}
\label{eq:LIFMathModel}
\tau_m \frac{\text{d}V}{\text{d}t} = V_{reset} - V + R_m I.
\end{equation}
where $\tau_m$ represents the membrane time constant, $V_{reset}$ is the resting potential, and $R_m$ and $I$ are the cell membrane's impedance and the input current, respectively.

The LIF neuron model provides a simplified representation of biological neurons, emphasizing essential features like membrane potential leakage, accumulation, and excitation. While its biological accuracy is limited, its simplicity makes it suitable for computational simulations.

\subsubsection{Other Models.}
The Hodgkin-Huxley (H-H) model, proposed by Hodgkin and Huxley in 1952~\cite{hodgkin1952HHModel}, offers a highly precise approximation of the principles governing biological neuron action potentials, earning them the Nobel Prize in Physiology or Medicine in 1963. Although the H-H model has high biological fidelity, it is computationally intensive. Other models, such as the Adaptive Exponential Integrate-and-Fire (aEIF) model~\cite{brette2005adaptive} and the Izhikevich model~\cite{izhikevich2003simple}, strive to balance biological fidelity and computational simplicity.

\subsection{Computational Unit Of SNNs}
In contemporary SNNs, neuron models predominantly rely on the LIF model. While the mathematical formulation of the LIF model involves a time-dependent differential equation, actual computer computations discretize this process for approximation.
\begin{equation}
    H[t] = V[t-1] + \frac{1}{\tau}(X[t] - (V[t-1]-V_{reset}))
\end{equation}
\begin{equation}
    S[t] = \Theta(H[t] - V_{threshold})
\end{equation}
\begin{equation}
    V[t] = H[t](1-S[t]) + V_{reset}S[t]
\end{equation}
\begin{equation}
    \Theta(x) = 
    \begin{cases} 
      0 & \text{if } x < 0 \\
      1 & \text{if } x \geq 0 
    \end{cases}
\end{equation}
where $\tau$ represents the membrane time constant, $V_{reset}$ is the resting potential, $V_{threshold}$ is the threshold voltage, and $X[t]$, $H[t]$, $V[t]$ represent the input current, the membrane potential before and after spiking firing at time step $t$, respectively. The specific implementation of LIF neurons can vary~\cite{fang2023spikingjelly}.

\tikzstyle{my-box}=[
 rectangle,
 draw=hidden-draw,
 rounded corners,
 text opacity=1,
 minimum height=1.5em,
 minimum width=5em,
 inner sep=2pt,
 align=center,
 fill opacity=.5,
 ]
 \tikzstyle{leaf}=[my-box, minimum height=1.5em,
 fill=hidden-orange!60, text=black, align=left,font=\scriptsize,
 inner xsep=2pt,
 inner ysep=4pt,
 ]
 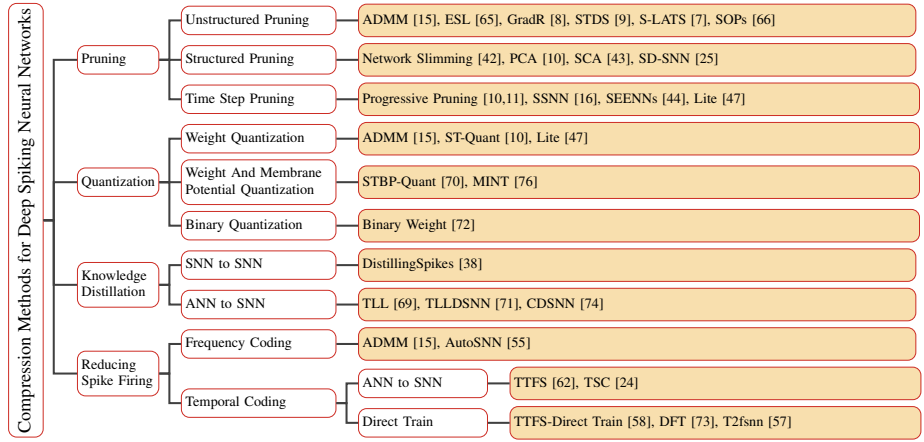
\begin{figure*}[h]
	\centering
	\resizebox{\textwidth}{!}{
		\begin{forest}
			forked edges,
			for tree={
				grow=east,
				reversed=true,
				anchor=base west,
				parent anchor=east,
				child anchor=west,
				base=left,
				font=\small,
				rectangle,
				draw=hidden-draw,
				rounded corners,
				align=left,
				minimum width=4em,
				edge+={darkgray, line width=1pt},
				s sep=3pt,
				inner xsep=2pt,
				inner ysep=3pt,
				ver/.style={rotate=90, child anchor=north, parent anchor=south, anchor=center},
			},
			where level=1{text width=4em,font=\scriptsize}{},
			where level=2{text width=8em,font=\scriptsize}{},
			where level=3{text width=6.6em,font=\scriptsize}{},
			[
			Compression Methods for Deep Spiking Neural Networks, ver
			[
			Pruning 
			[
			Unstructured  Pruning 		
			[
			ADMM~\cite{deng2021admm}{,}
			ESL~\cite{shen2023esl}{,}
			GradR~\cite{chen2021GradR}{,}
                STDS~\cite{chen2022stds}{,}
                S-LATS~\cite{chen2023unifiedThresholdPrun}{,}
                SOPs~\cite{shi2023sopbased}
			, leaf, text width=30em
			]
			]
			[
			Structured  Pruning
			[
                Network Slimming~\cite{li2024bnGammaPenaltyTermSNN}{,}
			PCA~\cite{chowdhury2021PCASNN}{,}
                SCA~\cite{li2024sca}{,}
                SD-SNN~\cite{han2022SDSNN}
			, leaf, text width=30em
			]
			]
                [
                Time Step Pruning
                [
                Progressive Pruning~\cite{chowdhury2021PCASNN,chowdhury2022towards}{,}
                SSNN~\cite{ding2024shrinking}{,}
                SEENNs~\cite{li2024seenn}{,}
                Lite~\cite{liu2024lite}
                , leaf, text width=30.2em
                ]
                ]
			]
			[
			Quantization
			[
			Weight Quantization
			[
			ADMM~\cite{deng2021admm}{,}
                ST-Quant~\cite{chowdhury2021PCASNN}{,}
                Lite~\cite{liu2024lite}
			, leaf, text width=30.2em
			]
			]
                [
                Weight And Membrane \\Potential Quantization
                [
                STBP-Quant~\cite{tan2023IntegerSTBP}{,}
                MINT~\cite{yin2024mint}
                , leaf, text width=30.2em
                ]
                ]
                [
                Binary Quantization
                [
                Binary Weight~\cite{wang2020binaryWeights}
                , leaf, text width=30.2em
                ]
                ]
			]
                [
			Knowledge \\ Distillation
			[
			    SNN to SNN
			[
			DistillingSpikes~\cite{kushawaha2021distilling}
			, leaf, text width=30em
			]
			]
			[
		      ANN to SNN
			[
                TLL~\cite{takuya2021training}{,}
                TLLDSNN~\cite{tran2022training}{,}
                CDSNN~\cite{xu2023constructing}
                , leaf, text width=30em
			]
			]
			]
			[
			Reducing \\Spike Firing
			[
                Frequency Coding
                [
                ADMM~\cite{deng2021admm}{,}
                AutoSNN~\cite{na2022autosnn}
                , leaf, text width=30em
                ]
			]
                [
                Temporal Coding
                [
                ANN to SNN
                [
                TTFS~\cite{rueckauer2018TTFS}{,}
                TSC~\cite{han2020tsc}
                , leaf, text width=22em
                ]
                ]
                [
                Direct Train
                [
                TTFS-Direct Train~\cite{park2021ttfsTrain}{,}
                DFT~\cite{wei2023temporalDirecTrain}{,}
                T2fsnn~\cite{Park2020T2FSNNDS}
                , leaf, text width=22em
                ]
                ]
                ]
			]
			]
		\end{forest}}
	\caption{Taxonomy of Model Compression methods for Deep Spiking Neural Networks.}
	\label{categorization}
\end{figure*}

\section{Methods}
DSNNs share many commonalities with DANNs, and some research has focused on these commonalities and proposed a unified pruning architecture~\cite{chen2023unifiedThresholdPrun}. Concepts like pruning, quantization, and knowledge distillation, initially developed for DANNs, can be adapted for DSNNs. However, the unique spiking and temporal characteristics of DSNNs often necessitate modifications to these techniques to ensure optimal performance. Additionally, DSNNs can be compressed in unique ways, such as leveraging spike sparsity, temporal coding, and pruning of time steps.

\subsection{Pruning}
DSNNs need to be deployed on neuromorphic hardware~\cite{yin2022sata}, which differs significantly from general-purpose computers in terms of operational logic. As a result, the value of pruning must be reassessed. Both unstructured and structured pruning are important for DSNNs, contrary to the inefficiency of unstructured pruning in DANNs~\cite{ma2022unstructuredPruningNoEfficient}, despite the scarcity of related work~\cite{liu2015sparseCANNs}.

\subsubsection{Unstructured Pruning.}
Unstructured pruning involves pruning weights or neurons.

ADMM-method~\cite{deng2021admm} utilizes the Alternating Direction Method of Multipliers (ADMM) for connection pruning based on soft constraints, demonstrating effectiveness in reducing parameter memory space and baseline computational cost in DSNNs. It compresses connections, weight bit-widths, and spike frequencies simultaneously.

ESL~\cite{shen2023esl} begins with a sparse network generated using the Erdős–Rényi random graph model, dynamically pruning weak connections and generating new ones according to structural plasticity rules. This approach, which updates the connection mask every $Titter$ iterations and employs various growth methods, outperforms the ADMM-method.

GradR~\cite{chen2021GradR}, inspired by synapse formation and elimination in the nervous system, redefines synaptic parameters using $w=sReLU(\theta)$, where $s$ is determined at initialization and remains unchanged. It employs different training strategies for active and inactive connections, enhancing network exploration and recovery of dead neurons.

STDS~\cite{chen2022stds} improves on GradR by using nonlinear reparameterization with $w=sign(\theta)\cdot (|\theta|-d)_{+},d\geq 0$. When the size of the connection $w$ is below threshold $d$, it is considered a filopodium, and the equivalent weight is zero. In any state, the parameters have gradients, allowing transitions between positive and negative weights, exploring the network space more fully. Pruning speed is adjusted by regulating the change in $d$.

S-LATS~\cite{chen2023unifiedThresholdPrun}, a theoretical framework that reformulates soft threshold pruning as an implicit optimization problem and solves it using the Iterative Shrinkage-Thresholding Algorithm (ISTA), which is a classic method in the fields of sparse recovery and compressed sensing.It is proven that in the underlying optimization problem, the L1 coefficient is jointly determined by the threshold and the learning rate, allowing any threshold tuning strategy to be interpreted as a scheme for adjusting the L1 penalty.Through in-depth research on threshold scheduling based on the framework, an optimal threshold scheduler is derived, which maintains a stable L1 regularization coefficient, thereby providing a time-invariant objective function from an optimization perspective.A new family of pruning algorithms is proposed, including pruning during training, early pruning, and pruning at initialization, and get the best result both in DANNs and DSNNs.

SOPs-method~\cite{shi2023sopbased} introduces an synaptic operation (SOP) metric (this paper define SOP as the operations performed when a spike passes through a synapse) to quantify power consumption in SNNs and uses an energy penalty term for energy-constrained unstructured weight and neuron pruning, maximizing efficiency through sparsity. During training, a binary mask $m$ is reparameterized by $\alpha$ as $m=H(\alpha)$ and approximated with a scaled sigmoid function as $\lim_{\beta \rightarrow \infty}\sigma(\alpha;\beta)=\frac{1}{1+e^{-\beta\alpha}}$. Gradual scheduling methods adjust the scale factor $\beta$, to achieve different compressing rate.

\subsubsection{Structured Pruning.}
Structured Pruning focuses on channel pruning for deep convolutional spiking neural networks. 

Just the migration from DANNs Networks Slimming~\cite{liu2017bnGammaPenaltyTermANN}, DSNNs Networks Slimming~\cite{li2024bnGammaPenaltyTermSNN} penalizes the scaling factors of the Batch-Normal layers, pruning channels with smaller scaling factors. However, due to the thresholding nature of DSNNs, the pruning is not thorough enough, achieving only 60\% channel pruning while maintaining accuracy without much loss.

Principal Component Analysis(PCA) based pruning~\cite{chowdhury2021PCASNN} targets redundant filters, a method also used in ANNs ~\cite{garg2019pcaCNN}, adapted for SNNs. The pruning method is modified by using the principal component analysis of neurons' average cumulative membrane potentials to determine significant spatial dimensions for structured pruning. This step results in a 10-14 fold reduction in model size.

The network pruning framework based on Spike Channel Activity(SCA)~\cite{li2024sca} is inspired by synaptic plasticity mechanisms. During training, channels with lower average spike firing frequency are pruned, and convolutional kernels are dynamically adjusted based on the gradients of the Batch-Normal layers scaling factors, regenerating those with larger gradients which is same to before work~\cite{evci2020rigging}.

Adaptive structural development of SNN(SD-SNN)~\cite{han2022SDSNN} is based on synaptic constraints inspired by dendritic spine plasticity. Synaptic constraints detect and remove significant redundancy in SNNs, and synapse regeneration effectively prevents and repairs excessive pruning. During the learning process, the neuron pruning rate and synapse regeneration rate are adaptively adjusted, leading to a stable SNN structure.

\subsubsection{Time Step Pruning.}
Time step pruning is unique to SNNs due to their temporal dimension in data input.

% FIGURE TODO(about how to multi and direct coding by first conv layer)

In algorithms transfer DANNs to DSNNs, a high number of time steps is needed to accurately capture the values of the ReLU activation function, requiring 20-50 steps or more. This is unacceptable for SNNs in practical applications, as it introduces step times computational and memory burdens~\cite{Bhattacharjee2023AreST}.

The main challenge of time step pruning is that when the number of time steps is too low, neurons in the latter parts of the model cannot fire, leading to vanishing gradients. This means back-propagation cannot occur, and the model cannot be trained.

A simple idea is progressive pruning~\cite{chowdhury2021PCASNN,chowdhury2022towards}. First, train the model with a long time step to ensure a high spike rate in the final layer. Then, prune the time steps while the model can still produce spikes in the final layer even with fewer time steps, allowing back-propagation to proceed normally. Gradually reduce the time steps until reaching an extremely low number, potentially achieving good results even with a single time step. The test dataset used for this approach is a standard image dataset.

Shrinking SNN (SSNN)~\cite{ding2024shrinking} addresses the issue of disappearing spikes by modifying the model structure, enabling the model to derive loss from the spikes of intermediate neurons, back-propagate, and train normally. This method has shown good results but is currently limited to the DVS dataset, demonstrating its compression capabilities on datasets with temporal sequences. It may also be extendable to static datasets.

Spiking Early-Exit Neural Networks (SEENNs)~\cite{li2024seenn} focus on balancing efficiency and accuracy. This method treats the time step as a variable and filters uncertain predictions through a confidence threshold. For uncertain predictions, reinforcement learning is used to determine the appropriate number of time steps, achieving automatic time step adjustment. On the CIFAR-10 dataset, this method achieved an average of nearly one time step while maintaining higher accuracy compared to other methods with longer time step.

Lite~\cite{liu2024lite} thoroughly explores the optimal number of time steps for each layer through neural architecture search, characterizes the energy consumption through the metric of total synaptic operations, and achieves a good balance between energy efficiency and accuracy by controlling the pruning strength with a penalty factor.

\subsection{Quantization}
Quantization transforms large numerical data into a discrete set of values, aiming to minimize bit usage while maintaining computational precision. In DSNNs, quantization includes membrane potential quantization alongside weights, without activation quantization. This study categorizes quantization efforts into weight quantization and weight and membrane potential quantization, along with specialized binary quantization.
\subsubsection{Weight Quantization.}
ADMM-Quant~\cite{deng2021admm} employs the ADMM to enforce quantization constraints on weights, albeit with a limitation to weight-only uniform quantization.

ST-Quant~\cite{chowdhury2021PCASNN} utilizes K-means clustering-based weight sharing quantization techniques to further compress the model, akin to the approach taken by ~\cite{han2015deepcompress} for weight quantization.

Lite~\cite{liu2024lite} adopts Neural Architecture Search (NAS) to automatically discover suitable mixed-precision bit-widths, enabling different layers to adopt varying levels of quantization. This is integrated within a unified framework that concurrently searches for optimal bit-widths, time steps, and network architectures, guided by a penalty term incorporated into the loss function for joint training.
\subsubsection{Weight And Membrane Potential Quantization.}
STBP-Quant~\cite{tan2023IntegerSTBP} transforms high-precision floating-point computations in existing direct training algorithms to low-bitwidth integer operations, introducing Integer-STBP, an algorithm that facilitates training and inference of SNNs using solely integer arithmetic. This enables implementation on low-power edge devices for online learning and inference.

MINT~\cite{yin2024mint}, building upon STBP-Quant, eliminates the requirement for multipliers in conventional uniform quantization during inference by sharing scaling factors between weights and membrane potentials. Scaling factors are retained during training to ensure competitive accuracy, while their necessity is obviated during inference through shared scaling, thus removing 32-bit multipliers in hardware. By quantizing memory-intensive membrane potentials to extremely low precision (2-bits), significant reductions in memory usage are achieved.

\subsubsection{Binary Quantization.}
A novel weight-threshold balancing transformation~\cite{wang2020binaryWeights} is proposed, adjusting the threshold of spiking neurons to convert high-precision weights into binary values (-1 or 1), effectively yielding binary SNNs. This drastically reduces weight memory requirements, albeit accompanied by increased neuron threshold storage (originally all are same).

\subsection{Knowledge Distillation}

Knowledge distillation (KD) is a commonly used model compression method, which uses a large teacher model to guide the training of a small student model. Specifically, knowledge distillation takes the knowledge of the teacher model as a supervision signal during the training process, so that the student model not only learns from the data, but also receives guidance from the teacher model, thereby achieving better performance.

\subsubsection{Knowledge distillation from SNN to SNN.}
As the scale of SNNs continues to expand, its demand for storage and computing resources also gradually increases, hindering its application in real life. To address this issue, Kushawaha et al. \cite{kushawaha2021distilling} proposed the first knowledge distillation method specially designed for SNNs to minimize loss of accuracy. They use the spiking activation tensors of the teacher and student models to simultaneously calculate the full loss and the sliding window loss as new loss functions. Then they froze the weights of the teacher SNN and trained the student SNN. Moreover, they also introduced a multi-stage distillation procedure to further improve the performance of student SNN. Experiments on three standard image classification datasets show that their method improves the performance of student SNN.

\subsubsection{Knowledge distillation from ANN to SNN.}
Takuya et al. \cite{takuya2021training} proposed to use knowledge distillation of KL divergence to train low-latency SNN. They first utilized a large ANN as the teacher model to train a small ANN, and then converted the small ANN into a SNN. Finally, they distill knowledge from the large teacher ANN into the SNN and use approximate gradients to solve the problem of non-differentiable spikes. The experimental results on the CIFAR-100 dataset show that they achieved the lowest inference latency while maintaining accuracy.

Tran et al. \cite{tran2022training} proposed a training technique to convert ANNs to SNNs which is able to learn more hidden information by using knowledge distillation. They combined knowledge distillation and batch normalization through time (BNTT) to improve the performance of converted SNNs and reduce its power consumption. Experiments on Tiny-ImageNet, CIFAR-10 and CIFAR-100 show that their method successfully improves accuracy and reduces inference latency.

Xu et al. \cite{xu2023constructing} proposed a knowledge distillation training method combining ANN and SNN. They combined spike coding and joint loss function to solve the problem of non-differentiable SNN spikes. They proposed two knowledge distillation methods: response-based KD and feature-based KD, which extract knowledge from the last layer output of the teacher model and some intermediate layers of the teacher model respectively. Comprehensive experimental results demonstrate the effectiveness of their approach.

\subsection{Reducing Firing Rate}
In neuromorphic hardware, a common metric for characterizing the energy consumption is synaptic operations~\cite{davies2018loihi}, and reducing the number of spikes can effectively decrease synaptic operations. Frequency coding is a commonly used encoding method, favored for its simplicity and efficiency, which is conducive to training; however, it typically requires a higher number of spikes to achieve satisfactory results under this encoding. Temporal coding, on the other hand, conveys the same information with fewer spikes, allowing at most a single spike per neuron in the neural network, effectively reducing the number of spikes during inference. The trade-off is longer time steps, which are typically used in the transition from ANN-SNN algorithms.Due to space constraints, we will only discuss the Temporal coding of ANN-SNN. Direct training temporal coding can be seen here~\cite{wei2023temporalDirecTrain,park2021ttfsTrain,Park2020T2FSNNDS}.
\subsubsection{Frequency Coding.}
ADMM-method~\cite{deng2021admm} reduces spike firing frequency during training by imposing a penalty factor on the membrane potential.
AutoSNN~\cite{na2022autosnn} is a direct training algorithm that adopts a one-shot weight-sharing approach based on an evolutionary algorithm to automatically search for and discover energy-efficient SNN architectures suitable for specific tasks. The article mentions that the global average pooling layer can reduce the energy efficiency of SNNs, so the maximum pooling layer is recommended.
\subsubsection{Temporal Coding.}
Time-to-First-Spike(TTFS) coding~\cite{rueckauer2018TTFS} approximates the real-valued ReLU activation in DANNs to the delay of the first spike in the corresponding spike sequence in DSNNs, requiring at most one spike per activation. However, the cost is a longer time step, and the memory access and computational costs associated with long time steps remain high, as peak neurons need to track synapses and receive the first peak from synapses after each time step.

Temporal-SwitchCoding (TSC), and the corresponding TSC spiking neuron model~\cite{han2020tsc}, is presented better then TTFS. Each pixel of the input image is presented through two spikes, with the time interval between the two spikes being proportional to the pixel intensity. Throughout the inference process, each synapse performs at most two memory accesses and two addition operations, significantly improving the energy efficiency of SNNs.

\section{Future Directions}
\subsubsection{Efficient Neuron Model.}
There are models that increase accuracy with more parameters and complex structures~\cite{hao2023LIFMHmodel,yao2022glif}, as well as models that precisely model biological neurons. However, there is limited research on developing more efficient neuron models specifically for deep learning~\cite{moser2024viewLIFasQuant}. Two potential paths exist: one involves modeling neurons more finely but with fewer neurons in shallow networks, and the other involves modeling neurons more coarsely with a larger number of neurons in deep networks. Precise modeling of biological neurons requires substantial computational resources but offers greater representational capacity. By accurately simulating biological neurons, we can achieve complex network functions with fewer neurons~\cite{lechner2020neural,beniaguev2021single,Zhao2024MetaWormAI}, which may present an energy efficiency advantage.

Balancing biological fidelity with computational efficiency presents a trade-off. Exploring how to abstract better neuron characteristics for modeling is a key problem that warrants further investigation to realize efficient neural networks.

\subsubsection{Unified Compression Architecture.}
DANNs and DSNNs share commonalities, and their similarities may represent the fundamental characteristics of neural networks. Methods applicable only to specific types of neural networks might not address the essence of neural networks. For instance, a unified framework for soft threshold pruning has been proposed, revealing the characteristics of neural network soft threshold pruning~\cite{chen2023unifiedThresholdPrun}. Similarly, the PCA method's analysis of convolutional kernel similarity followed by pruning demonstrates the effectiveness of pruning similar components in neural networks~\cite{garg2019pcaCNN,chowdhury2021PCASNN}. The difficulties encountered in regularization pruning of Batch-Normal layer parameters suggest that a unified solution is needed~\cite{li2024bnGammaPenaltyTermSNN,liu2017bnGammaPenaltyTermANN}.

Key issues in DSNNs compression include: What are the differences between DANNs and DSNNs during compression? Which DANNs methods can be transferred to DSNNs? If there are differences, how should they be modified? Is there a unified method?

\subsubsection{Efficient Specialization Program.}
Are there more efficient encoding schemes? Time encoding requires long time steps, and frequency encoding requires higher firing frequencies. Fully utilizing the time and frequency information of spikes is an important issue. Existing work combines time and frequency, but can we go further? Moreover, we may need more specialized model structures and further research on them, beyond just minor modifications to VGG and ResNet like VGGSNN and Sew-ResNet~\cite{ma2024TeCoS-LVM,lechner2020neural,hasani2021liquid}.
\subsubsection{Co-Design Compression With Hardware.}
The requirement for DSNNs to be deployed on neuromorphic hardware to gain advantages over DANNs affects the direction of DSNNs optimization. A variety of neuromorphic hardware platforms~\cite{davies2018loihi,pei2019tianjic,benjamin2014neurogrid,painkras2013spinnaker,schemmel2010wafer} may necessitate hardware-related collaborative compression work.

\subsubsection{Disclosure of Interests.} The authors have no competing interests to declare that are relevant to the content of this article.

\bibliographystyle{splncs04}
\bibliography{mybibliography}

\end{document}